\documentclass{article}

\usepackage{microtype}
\usepackage{graphicx}
\usepackage{booktabs} 
\usepackage{amsmath}
\usepackage{amsthm}
\usepackage{amsfonts}
\usepackage{graphicx}
\usepackage[ruled , linesnumbered]{algorithm2e}
\usepackage{float}
\usepackage{subcaption}
\usepackage{booktabs}
\usepackage{natbib}
\usepackage[table,xcdraw]{xcolor}
\usepackage{authblk}

\definecolor{colorMNIST}{HTML}{228833}
\definecolor{colorStack}{HTML}{EE6677}

\newcommand{\sref}[1]{{Sec}~\ref{#1}}
\newcommand{\fref}[1]{{Fig}~\ref{#1}}
\newcommand{\eref}[1]{{Eq}~\ref{#1}}
\newcommand{\tref}[1]{{Table}~\ref{#1}}
\newcommand{\aref}[1]{{Algorithm}~\ref{#1}}

\usepackage{hyperref}

\title{Communication-Efficient Federated Learning via Optimal Client Sampling}

\author[1]{M\'onica Ribero \footnote{ corresponding author: \texttt{ mribero@utexas.edu }} }
\author[1]{Haris Vikalo}
\affil[1]{Department of Electrical and Computer Engineering, University of Texas at Austin, Texas, USA}

\begin{document}

\maketitle

\begin{abstract}
Federated learning (FL) ameliorates privacy concerns in settings where a central server coordinates learning from data distributed across many clients. The clients train locally and communicate the models they learn to the server; aggregation of local models requires frequent communication of large amounts of information between the clients and the central server. We propose a novel, simple and efficient way of updating the central model in communication-constrained settings based on collecting models from clients with informative updates and estimating local updates that were not communicated.
In particular, modeling the progression of model's weights by an Ornstein-Uhlenbeck process allows us to derive an optimal sampling strategy for selecting a subset of clients with significant weight updates. The central server collects updated local models from only the selected clients and combines them with estimated model updates of the clients that were not selected for communication. We test this policy on a synthetic dataset for logistic regression and two FL benchmarks, namely, a classification task on EMNIST and a realistic language modeling task using the Shakespeare dataset. The results demonstrate that the proposed framework provides significant reduction in communication while maintaining competitive or achieving superior performance compared to a baseline. Our method represents a new line of strategies for communication-efficient FL that is orthogonal to the existing user-local methods such as quantization or sparsification, thus complementing rather than aiming to replace those existing methods.

\end{abstract}

\section{Introduction}
\label{sec:intro}

Federated learning (FL) is a privacy-preserving framework for training machine learning models in settings where the data is distributed across many clients. Such settings are common in applications that involve mobile devices \citep{mcmahan17a}, automated vehicles, and Internet-of-Things (IoT) systems, as well as in cross-silo applications including healthcare \citep{ribero2020federating} and banking. In \texttt{FedAvg}, the baseline FL procedure proposed in \citep{mcmahan17a} (included in the supplementary material as \aref{alg:FedAvg}), a server distributes an initial model to clients who independently update the model using their local training data. These updates are aggregated by the server
which broadcasts a new global model to the clients and selects a subset of them to start a new round of local training; the procedure is repeated until convergence. Since clients communicate only their models to the server, FL offers data security that can be further strengthened using privacy mechanisms including those that provide differential privacy guarantees \citep{abadi2016deep, wu2017bolt, mcmahan2018general}. 

The number of clients in FL systems may be in the order of millions, and the models that they locally train could be rather large; for example, VGG-16, the widely known neural network for image recognition has 138M parameters \citep{simonyan2014very}, weighing 526MB when represented by 32 bits. 
Moreover, many settings where FL is applied are highly dynamic (e.g., mobile devices, IoT), with new users joining at any moment and old users continuing to generate new data. Such settings may lead to a large number of training rounds and clients' model uploads, even though the contributions of some locally updated models to the global one may be rather limited. Since transmitting large models requires considerable communication resources, it is desirable to reduce the amount of information that has to be collected by the server. This is being explored in a promising line of work focused on reducing each client's communication budget by compressing the ML model through strategies such as quantization and sparsification \citep{tang2019doublesqueeze,konevcny2016federated,suresh2017distributed,konevcny2018randomized,alistarh2017qsgd,horvath2019natural}. 

In the existing FL systems, the number of clients participating in each round of updates (and, therefore, the required communication budget) is typically fixed. Yet the contributions of many clients in any given round may have limited impact, especially near convergence. Following this intuition, we propose a novel approach to reducing communication in FL by identifying and transmitting only the client updates that are deemed informative.\footnote{Note that this approach is orthogonal to the aforementioned compression strategies, and that the two may in principle be complemented (left to future work).} In particular, we model the progression of users' vector of weights during stochastic gradient descent (SGD) as a multidimensional stochastic process, and let each user decide whether or not to send an update to the server based on how informative is the observed segment of a sample path (e.g., how far is the process from its steady-state). Specifically, we rely on the Ornstein-Uhlenbeck process (OU) -- a continuous stochastic process parameterized by the mean and covariance functions -- and interpret weights in SGD iterations as the points obtained by discretizing a sample path of the underlying OU process. 
Relying on this connection, we borrow techniques for optimal sampling of OU processes and adapt them to the problem of optimal client sampling. The optimal strategy turns out to be a simple threshold on the update's norm and can thus be efficiently implemented at the client side. 

We develop a dynamic strategy for selecting the threshold value; the threshold varies adaptively during the training process. In particular, after a round of training on local data, the clients share their updates' norm with the server; the server computes a threshold and broadcasts it to the clients who in turn rely on the received threshold to decide whether or not to send their updates to the server. Updates of the clients who chose not to communicate are predicted using the server's estimate of the parameters of the underlying OU process. Heuristics that discard small model updates in distributed learning have previously been proposed  \cite{hsieh2017gaia, chen2018lag, singh2019sparq}; however, in this prior work the missing models are replaced with either historic values or simply ignored. In contrast, we rely on our OU model assumptions to estimate missing model updates, and demonstrate that such a non-trivial update boosts performance of the FL scheme. 

Efficacy of the proposed formulation is demonstrated in practical settings where we show that it may reduce communication during an FL training loop up to 50 \% while achieving the same rate of convergence and competitive (or better) model accuracy. In particular, we test our methods on a logistic regression with synthetic data, and two FL benchmark tests: a 62 character classification task on a real federated dataset EMNIST \cite{cohen2017emnist} using a convolutional neural network architecture, and on the Shakespeare dataset \cite{mcmahan17a} where we trained a model for a character prediction task. 

\begin{figure*}
    \centering
\begin{subfigure}{0.23\textwidth}
    \includegraphics[width = \textwidth]{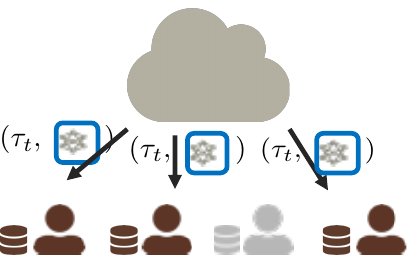}
    \caption{Server selects $N$ clients, and broadcasts the current global model and threshold.}
    \label{fig:diag1}
\end{subfigure}
\hfill
\begin{subfigure}{0.36\textwidth}
    \includegraphics[width = \textwidth]{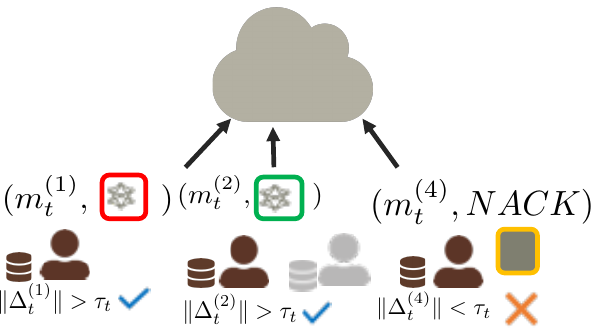}
    \caption{Client $k$ locally updates her model and computes $\Delta_t^{(k)}$. If clients meet the threshold, they send back the update and its norm (in the illustration, clients 1 and 2). Otherwise, clients only send the norm of their update (client 4).}
    \label{fig:diag2}
\end{subfigure}
\hfill
\begin{subfigure}{0.35\textwidth}
\centering
    \includegraphics[width = 0.6\textwidth]{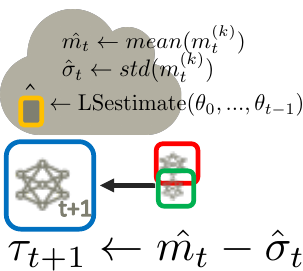}
    \caption{In round $t$, server: (i) computes the mean and standard deviation of the received norms of the updates; (ii) estimates missing clients updates via (\ref{eq:ou_estimator}); (iii) updates the global model and computes the threshold for round $t+1$.}
    \label{fig:diag3}
\end{subfigure}
    \caption{One round $(t)$ of communication efficient federated learning system through client sampling and estimation}
    \label{fig:diagram}
\end{figure*}

We summarize our proposed scheme in \fref{fig:diagram}. Our approach requires two major modifications of the original Federated Averaging algorithm (\texttt{FedAvg}, included for completeness in \sref{sec:fedavg} as Algorithm~\ref{alg:FedAvg}). As in the original \texttt{FedAvg} \cite{mcmahan17a}, the server selects $N$ clients and broadcasts the current model, but now includes a threshold $\tau_t$ (\fref{fig:diag1}). The selected clients train for $E$ epochs and evaluate how much their update differs from the original model by computing the norm of the difference between the models. If this difference exceeds threshold $\tau_t$, the update is transmitted; otherwise, only the norm of the update is sent to the server (\fref{fig:diag2}). The server relies on the model history to estimate parameters of the underlying OU process, and uses them to predict missing clients updates (rather than zeroing or ignoring these clients updates). This is described in details in \sref{sec:formulation}. By combining received and estimated updates, the server generates a new model. Finally, the server uses the collected local model norms to produce a new threshold for the next round; the described procedure repeats until a stopping criterion is met.

\section{Background and Related Work}
\label{sec:background}

\subsection{Federated Learning}

The FL algorithm introduced in \cite{mcmahan17a}, \texttt{FedAvg}, requires a random subset of clients to send their updates to the server after having trained locally for $E$ epochs on mini-batches of size $B$. For more details, please see \aref{alg:FedAvg} in the supplementary document. In subsequent years, various aspects of FL systems have attracted a lot of attention including privacy and security \cite{abadi2016deep, geyer2017differentially, pmlr-v97-amin19a, bonawitz2016practical}, optimization \cite{reddi2020adaptive}, adversarial attacks \cite{bagdasaryan2018backdoor, bhagoji2018analyzing} and  personalization \cite{smith2017federated} -- see \cite{kairouz2019advances} for a comprehensive overview. Our focus is on communication challenges in FL. 

\subsection{Communication reduction strategies}

Highly relevant to our problem is the line of research on distributed estimation under communication constraints (see, e.g., \cite{boyd2011distributed,duchi2011dual,balcan2012distributed,zhang2013communication} and the references therein). Recently, work by  \citet{zhang2013information, braverman2016communication, han2018geometric, acharya2018inference} provided bounds on the minimax risk, i.e., the worst-case estimation error for a distributed system operating under a constraint on the maximum number of bits that remote nodes are allowed to transmit. Under the same constraint, \citet{barnes2019learning} leveraged Fisher information to derive lower bounds on the estimation error. 

Existing schemes for reducing communication overhead in FL systems typically perform compression on the client side and thus require additional computation for encoding and decoding. Deterministic approaches such as low rank approximation, sparsification, subsampling, and quantization \citep{konevcny2016federated, alistarh2017qsgd, horvath2019natural}, as well as randomized approaches including random rotations and stochastic rounding \citep{suresh2017distributed} and randomized approximation \citep{konevcny2018randomized}, can be used to reduce the communication while maintaining high accuracy. Note that these methods may be leveraged on the server side as well \citep{caldas2018expanding}. 

Orthogonal to compression-based methods, approaches that dismiss updates of some workers have been proposed in the distributed learning literature \cite{ hsieh2017gaia, chen2018lag,singh2019sparq}. \citet{hsieh2017gaia} propose to threshold significant updates where the significance is measured in terms of relative magnitude change; only those updates that exceed a threshold are communicated and averaged. A major drawback of this method is ignoring updates near convergence, thus causing stagnation in training. Our experiments demonstrate that such effects lead to unstable behavior, particularly in the case of heterogeneous data. Thresholding method in \cite{chen2018lag} explores settings that rely on a central server for coordination and proposes to replace the update of a client that did not communicate by the client's previous update. This approach requires the server to store for all clients their updates from the previous round. \cite{singh2019sparq} considers a similar approach for the fully decentralized setting, with a notable difference of setting the threshold according to a schedule $c_t = o(t)$ to guarantee convergence. These last two approaches are challenging to implement in FL settings because the number of clients in FL can be on the order of millions, and it is thus very likely that in each training round new clients are sampled. Therefore, updates of the clients that did not communicate should be replaced by the latest model they received, likely slowing down the training process. Nevertheless, this direction is worth pursuing since reducing the number of clients that send updates to a server may significantly reduce the amount of communication, as demonstrated by the heuristics which impose limits on the uploading times of updates \citep{Nishio2019client}. On a different vein, \citet{cho2020client} theoretically analyzes the effect of biased sampling on convergence. 

Another approach to learning under resource constraints is focused on reducing the overall model complexity, e.g., by bounding the model size \citep{lin2016fixed,oktay2019model}, pruning \citep{han2015deep, aghasi2016net, dong2017learning, zhu2017prune}, or restricting weights to be binary \citep{courbariaux2015binaryconnect}. 
Adaptation of these methods to FL and their analysis in such context are open research questions.


\subsection{Ornstein-Uhlenbeck Process}\label{sec:ouProcess}

In this section we provide a brief background on the Ornstein-Uhlenbeck process and overview an efficient procedure for the estimation of the process parameters. These concepts are at the core of our proposed approach to estimating client updates that due to thresholding are not communicated to the server in FL systems.

\paragraph{Definition.}
The Ornstein-Uhlenbeck process (OU) is a stationary Gauss-Markov process which, over time, drifts towards its mean function. Unlike the Wiener process whose drift is constant, the drift of the OU process depends on how far is its current realization from the mean. The OU processes have been extensively studied in a wide range of fields including physics \citep{lemons2002introduction}, finance \citep{stein1991stock,evans1994modelling,siu2015optimal}, and biology \citep{ricciardi1979ornstein, rohlfs2014modeling}, to name a few. Formally, the OU process $\{x_t \}_t$ is described by the stochastic differential equation 
  \begin{equation} \label{eq:SDE}
     d\theta_t = \lambda(\mu - \theta_t)dt+\sigma dW_t,
 \end{equation}
where $W_t$ denotes the standard Wiener process. Expression (\ref{eq:SDE}) specifies the process that is drifting towards $\mu$ with velocity $\lambda$, and has volatility driven by a Brownian motion with variance $\sigma$.

\paragraph{Estimating parameters of the OU process.} Various techniques for estimating parameters of the OU process from the observations of its sample path have been proposed in literature, including least-squares, maximum likelihood \citep{liptser2013statistics} and Jackknife method \citep{shao2012jackknife}. For convenience, we here summarize the computationally efficient least-squares solution. First, note that by discretizing the continuous OU process we obtain
 \begin{equation}
     \theta_{t+1} = e^{-\lambda\Delta t}\theta_t + (1-e^{-\lambda\Delta t})\mu + \sigma \sqrt{\frac{1-e^{-\lambda \Delta t}}{2\lambda}}\Delta W_t, \label{eq:discrete_OU}
\end{equation}
where $\Delta t$ denotes the discretization (sampling) period and $\Delta W_t$ are i.i.d. increments of the Wiener process. This leads to a linear measurement model
 \begin{equation}
 \label{eq:LS}
     \theta_{t+1} = a\theta_t + b + \epsilon_t
 \end{equation}
where $\epsilon_t$ denotes i.i.d. noise and where
\begin{equation*}
      a= e^{- \lambda \Delta t}, \;  b= \mu(1-e^{\lambda \Delta t}), \; std(\epsilon_t) = \sigma \sqrt{\frac{(1-e^{2 \lambda \Delta})}{2\lambda}}.
\end{equation*}



To enable efficient online (i.e., recursive) estimation of the relevant process parameters, let us define 
\begin{eqnarray}
\nonumber S_{x, t} = \sum_{i=1}^t \theta_{i-1}, & S_{y,t} = \sum_{i=1}^t \theta_{i}, \\
\nonumber S_{xx, t} = \sum_{i=1}^t \theta^2_{i-1}, & S_{yy, t} = \sum_{i=1}^t \theta_{i}^2, \\
\label{eq:rolling_sums} S_{xy, t} = \sum_{i=1}^t \theta_{i-1}\theta_{i}, & 
\end{eqnarray}
where $\theta_1, \theta_2, \dots \theta_t$ denote samples of the OU process. It is straightforward to show that the least-square estimates of $a_t$, $b_t$ and $std(\epsilon_t)$ given $\theta_1, \theta_2, \dots \theta_t$ can be found as

\begin{align}
 \label{eq:estimate_a_b}    
 \hat{a}_t = \frac{tS_{xy, t} - S_{x, t}S_{y, t}}{tS_{xx, t} S_{xx, t}}, & \quad \hat{b}_t = \frac{S_{y, t}-\hat{a}_tS_{x, t}}{t}
\end{align}
and
\begin{equation*}
    \widehat{std(\epsilon_t)} = \sqrt{\frac{tS_{yy, t} - S_{y, t}^2 - \hat{a}_t(tS_{xy, t} - S_{x, t}S_{y, t}) }{t(t-1)}}.
\end{equation*}

Finally, the next value of the sample path, $\theta_{t+1}$, is predicted as

\begin{equation}
\label{eq:predict_next_step}
    \hat{\theta}_{t+1} = \hat{a}_t \theta_t + \hat{b}_t.
\end{equation}


\paragraph{Optimal sampling of the OU process.} Rate-constrained sampling of stochastic processes has been widely studied in literature, primarily in the context of communications and control \citep{imer2005optimal, bommannavar2008optimal,nayyar2013optimal,rabi2012adaptive,Nar2014SamplingMulti,sun2017remote,ornee2019sampling, guo2020optimal}.
In \citep{imer2005optimal, bommannavar2008optimal}, this problem is studied for random i.i.d. sequences and Gauss-Markov processes;  \citet{nayyar2013optimal} present a similar study in the scenario where the nodes of a sensor network collaboratively estimate environment while operating under energy constraints that limit the amount of information sensors can transmit to a central processor.  \citet{rabi2012adaptive} study linear diffusion processes and formulate sampling as a stopping time problem;  \citet{Nar2014SamplingMulti} extend these results to multidimensional problems.

The setting where samples are observed locally (by nodes/clients) but used for estimation only if communicated to the central processor is studied in \citep{sun2017remote, ornee2019sampling}. As shown there, thresholding the signal magnitude increment is an optimal sampling policy for estimating parameters of an OU process.  An extension of these results to a larger class of continuous Markov processes with regularity conditions is considered in \cite{guo2020optimal}.

\paragraph{Connection between OU processes and SGD.} Recently,  \citet{blanc2019implicit, wang2017asymptotic, li2018statistical,mandt2016variational} have studied SGD in various settings by relying on stochastic differential equation models.  \citet{mandt2016variational} model SGD as an OU process and leverage its properties to derive optimal model parameters.  \citet{wang2017asymptotic} investigates asymptotic behaviour of descent algorithms using stochastic process models. \citet{li2018statistical} show that SGD can be used for statistical inference and demonstrate that the average of SGD sequences can be approximated by an OU process. \citet{blanc2019implicit} show that when learning a neural network with SGD and independent label noise, the dynamics of weight updates can be interpreted as an OU process. This connection is formalized in the next section.

\section{Efficient Training via Client Sampling}

\label{sec:formulation}

Here we introduce the techniques from optimal process sampling to FL systems, with the goal of reducing communication and improving accuracy. In particular, we present a strategy where a client transmits its weights to the server only if the norm of the client's model update exceeds a pre-determined threshold, and provide a non-trivial estimator of the model updates that did not meet the communication threshold. 

For convenience, the notation, including definition of variables, is summarized in \tref{tab:notation}

\begin{table}[ht]
\centering
\begin{tabular}{ll}
\hline
Variable                                             & Definition                                                         \\ \hline
$N$                                                    & Number of clients to sample \\
                                                     & each round     \\
$T$                                                    & Total number of rounds                                             \\
$\tau_t$                                             & Threshold sent by the server to \\ 
& users at the beginning of round $t$ \\
$\theta_t$                                           & Global model at round $t$                                          \\
$\theta^{(k)}_{t+1}$                                 & Local model at the end of \\ &
round $t$ at client $k$                  \\
$\Delta^{(k)}_{t} : = \theta^{(k)}_{t+1} - \theta_t$ & Update for client $k$ at round $t$                                 \\
$m_t^{(k)}:=\| \Delta^{(k)}_{t}\|_2$                 & $\ell_2$- norm of client $k$ update at $t$                    \\ 
$m_t$                                                & Mean of updates norms at the \\
& end of round $t$                      \\
$s_t$                                                & Standard deviation of norms  \\
& at the end of round $t$  \\ \bottomrule
\end{tabular}
\caption{Notation used in the paper.}
\label{tab:notation}
\end{table}

\subsection{SGD as an OU process}

Consider the loss function $\mathcal{L}(\theta;X) = \sum_{i=1}^N \ell_i(\theta)$, where $X$ is a dataset with $N$ samples and $\ell_i(\theta)$ is the loss of point $x_i \in X$ for $i=1,\dots, N$. In gradient descent, $\mathcal{L}$ is minimized by evaluating in each iteration an approximation of the gradient using a mini-batch $\mathcal{S} \subseteq X$ of the data. In particular, $$\theta_{t+1} \gets \theta_t - \frac{\eta}{|\mathcal{S}|} \sum_{i \in \mathcal{S}}g_i(\theta).$$ 
 The following observations and assumptions are commonly encountered in literature (see, e.g., \citep{mandt2016variational}).
 
\textit{Observation 1:} The central limit theorem implies that $\frac{1}{|\mathcal{S}|} \sum_{i \in \mathcal{S}}g_i(\theta) \rightarrow \mathcal{N}(g(\theta), B(\theta)B(\theta)^T)$, where $g(\theta)$ denotes the full gradient and $B(\theta)B(\theta)^T$ is the corresponding covariance matrix.

\textit{Assumption 1:} When $\theta$ approaches a stationary value, $B(\theta) = B$ is constant \citep{mandt2016variational}. 
 
\textit{Assumption 2:} The iterates $\theta_t$ lie in a region where the loss can be approximated by a quadratic form $ \mathcal{L}(\theta) = \frac{1}{2} \theta^TA\theta$ (readily justified in the case of smooth loss functions), and the process reaches a quasi-stationary distribution around a local minimum. 
 

 
 
Predicated on the above, the discrete process
$$\Delta\theta = \theta_{t+1}-\theta_t \approx -\eta g(\theta) - \sqrt{\frac{\eta}{N}} B \mathcal{N}(0, \eta I)$$ 
can be interpreted as obtained by discretizing the OU process
$$d\theta_t = -g(\theta)dt + \sqrt{\frac{\eta}{N}}BdW_t = -A\theta_t dt + \sqrt{\frac{\eta}{N}}BdW_t$$
in the relative proximity of the steady state. In the supplementary material (\sref{sec:numerical_illustration}) we illustrate these arguments by showing plots of the sample paths of randomly selected weights in a convolutional neural network trained on the Fashion MNIST dataset \cite{xiao2017/online} using SGD with a constant learning rate. At each iteration, distribution of the weight increments is approximately Gaussian; as expected, the weights follow trajectories typical of OU sample paths.

\subsection{Optimal sampling of OU processes in FL settings}
The OU process sampling strategies discussed in \sref{sec:background} assume a network of nodes with unlimited access to the process, while the estimator is located on a central server. 
When a sampling frequency constraint (i.e., uplink bandwidth) limits the number of samples that could be collected by the server, the nodes should locally decide when to send an update. Guo et al. \cite{guo2020optimal} show that the optimal strategy minimizing Mean Squared Error (MSE)  is to sample at time $\tau= \inf\{ t\geq 0 : |\theta_t - \mathbb{E}[\theta_t | \theta_{0}] | >\gamma \}$, and that the optimal decoding policy is given by 
$\hat{\theta}_t = \mathbb{E}[\theta_t|\theta_{0}]$, for  $t \in [0, \tau)$. In particular, for the OU process in (\ref{eq:SDE}),

\begin{equation} \label{eq:ou_estimator}
    \hat{\theta}_t = e^{-\lambda \Delta t} \theta_0+ (1-e^{-\lambda \Delta t}) \mu.
\end{equation}

To establish a connection to FL, we recall the arguments from the previous subsection and note that in each round $t$ of an FL procedure, client $k$ ``observes" a partial sample path of an OU process that terminates in $\theta_t^{(k)}$ (i.e., the client records progression of its weights during local training); the sample path starts from point $\theta_t$ (i.e., model weights) broadcasted by the server at the beginning of the current training round. Let $\Delta^{(k)}_{t} : = \theta^{(k)}_t - \theta_t$ be the difference between a locally updated (by client $k$) and the previously broadcasted model. Invoking the above sampling optimality results, we propose to schedule transmission of updates if $\|\Delta^{(k)}_{t}\|_2$ exceeds a judiciously selected threshold. We formalize this modified client update procedure as Algorithm \ref{alg:client_update}.

If following the result of a thresholding test client $k$ decides not to communicate with the server, the server may estimate the client's update according to (\ref{eq:ou_estimator}). Since the process parameters $\lambda$ and $\mu$ are unknown, they need to be estimated; equivalently, we replace (\ref{eq:ou_estimator}) by (\ref{eq:predict_next_step}) where $\hat{a}_t$ and $\hat{b}_t$ are inferred using the previously aggregated $\theta_0,\dots, \theta_t$ available to the server (as described in \sref{sec:ouProcess}). By storing rolling sums of the previous models, defined in (\ref{eq:rolling_sums}), the server can recursively update $\hat{a}_t$ and $\hat{b}_t$ and thus efficiently evaluate $\hat{\theta}_{t+1}$. 
We formalize this procedure as Algorithm \ref{alg:server_estimate}.
\subsection{Adaptively selecting the threshold}
\label{sec:threshSel}

The policy of deciding whether or not to communicate based on comparing $\|\Delta^{(k)}_{t}\|_2$ to a threshold $\tau_t$ aims to reduce communication without incurring significant accuracy loss compared to a baseline. Since the norm of the gradients is expected to decrease as the training progresses, a fixed threshold may have detrimental effect on the learning process as we get close to a minimum. We empirically explore this point in  \sref{sec:fixed_threshold}; in particular, we propose a strategy for adaptive modification of $\tau_t$ based on the magnitudes of the updates of participating clients. At $t=0$, clients receive an initial threshold $\tau_0 = 0$, implying that everyone transmits in the first round. In the following rounds, clients transmit either their updates and the norm of their update $m_t^{(k)}$, or a NACK message along with the norm of their update.  At the end of round $t$, server estimates the mean of the norms of updates, $m_{t}$, their standard deviation, $s_{t}$, and computes the threshold for the next round, $\tau_{t+1} = m_t -s_t $. 


\subsection{A communication-efficient algorithm }

Summarizing the discussions, we formalize our framework for communication-efficient FL as \aref{alg:all_system}. In brief, $N$ clients are selected at the beginning of round $t$. The server broadcasts the model parameters $\theta_t$, and the selected clients locally performs SGD with mini-batches of size $B$ for $E$ epochs.
Following a comparison of the norm of the local model update to a threshold, each client \textit{locally} decides whether to communicate its updates or not, and transmits either the model updates $\theta^{(k)}_{t+1}$ or a negative-acknowledgement message, respectively. In both cases, the client sends its training data size $n_k$ to enable weighting by $w_k = \frac{n_k}{\sum_jn_j}$ (line 16). 
Each client also transmits to server the norm of its update (only one float, 32 bits, compared with full model), even if it does not transmit the actual update.  Finally, the server estimates $k^{th}$ client's parameters as
\begin{equation} \label{eq:estimation}
\hat{\theta}_{t+1}^{(k)} = \begin{cases}
\theta^{(k)}_{t+1}, &  \mbox{if received an update,} \\
\hat{a}_t \theta_t + \hat{b}_t, &  \mbox{ otherwise. }
\end{cases}
\end{equation}
    
Parameters $\hat{a}_t$ and $\hat{b}_t$ are estimated via the least-squares procedure in \sref{sec:ouProcess}. Note that the server's computationally cheap alternative to estimation is to reuse the client’s model from the previous round, i.e., to set
$\hat{\theta}_{t+1}^{(k)} = \theta_t$, or to simply ignore the client; as reported in the next section, these alternatives consistently underperform our proposed policy. 
Finally, the server computes a new model according to $\theta_{t+1} = \sum_{k=1}^N w_k\hat{\theta}_{t+1}^{(k)}$ 
(line 13 of the pseudo-code), updates the threshold and the rolling sums used for parameter estimation, and proceeds to the next round of training.

\begin{algorithm}[]
    \KwIn{$K$ clients, local minibatch size $B$, number of local epochs  $E$,  learning rate  $\eta$, threshold selection rule $R$ }
    \KwOut{ Global model $\theta$ }
    initialize $\theta_0$\;
    initialize prediction $\hat{\theta}_1$ and rolling sums $\mathcal{S} = [S_x, S_{xx}, S_y, S_{yy}, S_{xy}]$\ at server\;
          \For{$t=1,2,...$}{ 
             $C_t \gets$ random set of $N$ clients \;
             \For{ $k\in C_t \quad $ \texttt{in parallel clients}}{
                 $M_t^{(k)} \gets$ ClientUpdate($k,\theta_t, E,B,\tau_{t}$) \;
             }
            \For{$k \in C_t \quad$ \texttt{ Server $\quad $ }}{
                 $\hat{\theta}_{t+1}^{^{(k)}} \gets \text{ServerEstimate}(M_t^{(k)}, \hat{\theta}_{t+1}) $(\eref{eq:estimation}) \;
                } 
            \texttt{Server: } \;
             $\tau_{t+1} = \text{mean}(m_t^{(k)}- \text{std}(m_t^{(k)})$\;
             $\theta_{t+1} \gets \sum_{k=1}^K w_k \hat{\theta}_{t+1}^{(k)}$ \;
            Update $\mathcal{S}$ with $\theta_{t+1} \quad $ (\eref{eq:rolling_sums})\;
            Predict $\hat{\theta}_{t+2} \quad $ (\eref{eq:predict_next_step}) \;
             }
  \caption{\texttt{Communication-Efficient FedAvg}}
 \label{alg:all_system}
\end{algorithm}

\begin{algorithm}[ht]
    \KwIn{Intitial global model $\theta_t$, threshold $\tau_t$, batch size $B$, number of epochs $E$, total number of samples $n_k$ at client $k$}
    \KwOut{ Message to server $M_t^{(k)}$ }
    initialize $\theta_{t+1}^{(k)} \gets \theta_t$\;
     $\mathcal{B} \gets $ split data $\mathcal{P}_k$ into batches of size $B$ \;
         \For{$i =1,...,E$}{
             \For{$b\in \mathcal{B}$}{
                 $\theta_{t+1}^{(k)} \gets \theta_{t+1}^{(k)}-\eta \nabla \ell(\theta_{t+1}^{(k)},b)$ \;
           }
         }
         $\Delta_{t}^{(k)} = \theta^{(k)}_{t+1} -\theta^{(k)}_{t+1}$\;
        $ m_t^{k} = \|\Delta_{t}^{(k)}\|_2$\;
         \eIf{$ m_t >\tau_t$}{
             $M_t^{(k)} \gets (\Delta_{t}^{(k)}, n_i,m_t^{(k)})$\;}
             {
             $M_t^{(k)} \gets (NACK,n_i,m_t^{(k)})$ \;  
                 }
       \Return{$M_t^{(k)}$ to server} \;
 \caption{\texttt{ClientUpdate} at client $k$}
 \label{alg:client_update}
\end{algorithm}

\begin{algorithm}[ht]
    \KwIn{Client $k$ message at time $t$, $M_{t}^{(k)}$, estimated model $\hat{\theta}_{t+1}$ for time $t+1$}
    \KwOut{ Estimate $\theta_{t+1}^{(k)}$ of client $k$for next global model }
    \begin{equation} 
        \hat{\theta}_{t+1}^{(k)} = \begin{cases}
        \theta^{(k)}_{t+1}, &  \mbox{ received updates} \\
        \hat{\theta}_{t+1}, &  \mbox{ otherwise. }
        \end{cases}
    \end{equation}
       \Return{Estimate $\hat{\theta}_{t+1}^{(k)}$ for client $k$} \;
 \caption{\texttt{ServerEstimate} for client $k$}
 \label{alg:server_estimate}
\end{algorithm}

\subsection{Computational complexity}

Computational complexity on the client side is a major challenge in FL due to limited power and memory of users' devices. Our proposed framework does not contribute to the clients' computational burden since the only additional operations on the client side are: (1) computing the update norm, and (2) comparing the norm with the threshold. The former is often already computed in FL systems to enable clipping large updates or to bound gradient sensitivity and guarantee differential privacy; the latter is negligible. On the server side, our framework increases memory consumption linearly in the number of parameters (in order to store the five arrays in $\mathcal{S}$, see \aref{alg:all_system}). Moreover, an additional computational overhead is needed to predict missing model updates via recursive least squares. This step is performed independently for each weight, and only requires element-wise sums and multiplications; thus the computational cost increases only linearly in the number of parameters, and does not change asymptotically.

\section{Experiments}
\label{sec:experiments}

In this section we present a number of FL experiments that demonstrate the performance of our proposed algorithm on different datasets and for various settings and models. In particular, we benchmark the proposed client selection strategy on three different datasets (a synthetic dataset, EMNIST, and Shakespeare) with three different models: (1) logistic regression on the synthetic dataset; (2) a more sophisticated convolutional neural network applied to EMNIST with 62 categories \cite{cohen2017emnist}; and (3) a recurrent neural network for the next character prediction on the Shakespeare dataset \cite{mcmahan2018general}. 
Further details and additional experimental results can be found in the supplementary document.

\subsection{Federated Datasets}

Here we elaborate on the different datasets used in benchmarking and model validation experiments. For the federated experiments with homogeneous data and convex learning objective, we use a synthetic dataset and a logistic regression task. We generate this data by taking $10^4$ samples $X_i \in \mathbb{R}^{100} \sim \mathcal{N}(0,I_{100})$. Moreover, we generate $\beta\sim \mathcal{N}(0,I_{100})$ and, finally, set labels $y_i = round(X_i^T\beta)$. The samples are split evenly among $100$ clients. 

For more realistic tasks we rely on the EMNIST dataset \cite{cohen2017emnist}, a reprocessed version of the original MNIST dataset with 62 categories: each image is a character linked to its original writer, providing a non-i.i.d. natural distribution and allowing us to emulate an FL setting. This dataset consists of images attributed to 3843 users. For the task of interest, character recognition, we train and test a convolutional neural network (CNN).

To investigate a language modelling task under data heterogeneity, we use the Shakespeare dataset \cite{mcmahan17a} -- a language modelling dataset with 725 clients, each one a different speaking role in each play from the collective works of William Shakespeare. Each client's dataset is restricted to have at most 128 sentences, and is split into training and validation sets.  Following the previous work with this dataset \cite{reddi2020adaptive}, we use a build vocabulary with 86 characters contained in the text, and 4 characters representing padding, out-of-vocabulary, beginning, and end of line tokens. We use padding and truncation to enforce 20 word sentences, and represent them with index sequences corresponding to the vocabulary words, out of vocabulary words, beginning and end of sentences. Finally, we train a recurrent neural network (RNN) with just under 1M parameters for the next character prediction task. Further details on the model architectures can be found in the appendix in \sref{sec:app:architectures}.

\subsection{Results}

Details of the experiments on the aforementioned federated datasets with three different architectures are elaborated below. The main results are presented in this section, with further details, plots and tables found in the supplementary document.

\begin{figure*}[]
    \centering
    \includegraphics[width=\textwidth]{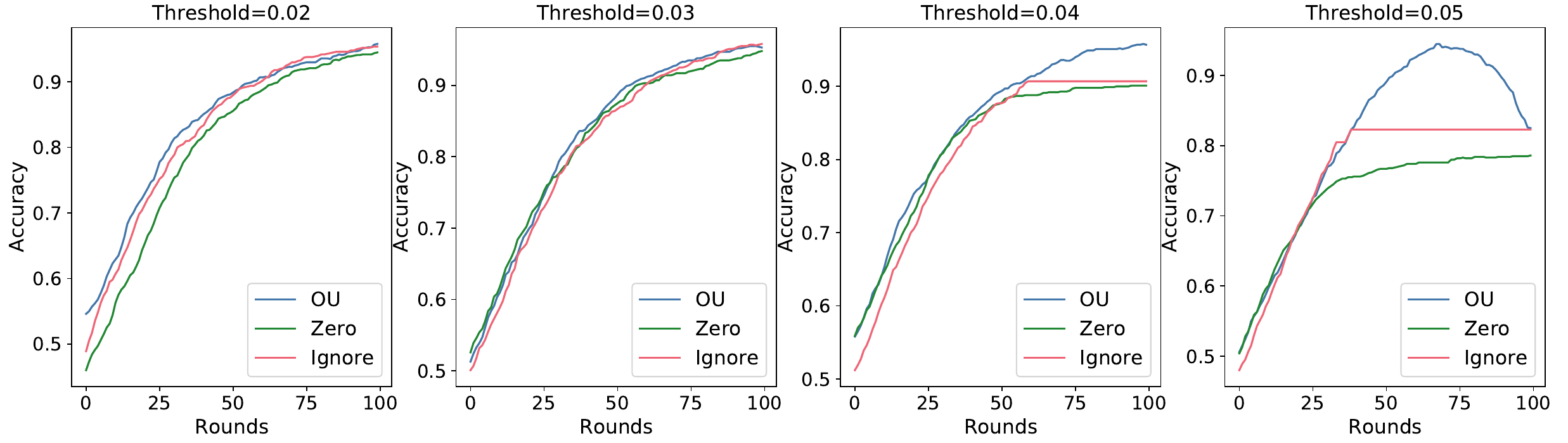}
    \caption{Accuracy of FedAvg with a fixed threshold communication strategy for varied values of the threshold. High thresholds prohibit convergence to an accurate model since all clients stop transmitting once their updates become smaller than the threshold.}
    \label{fig:fixed_threshold_accuracy}
\end{figure*}

\begin{figure*}[ht]
    \centering
    \includegraphics[width=\textwidth]{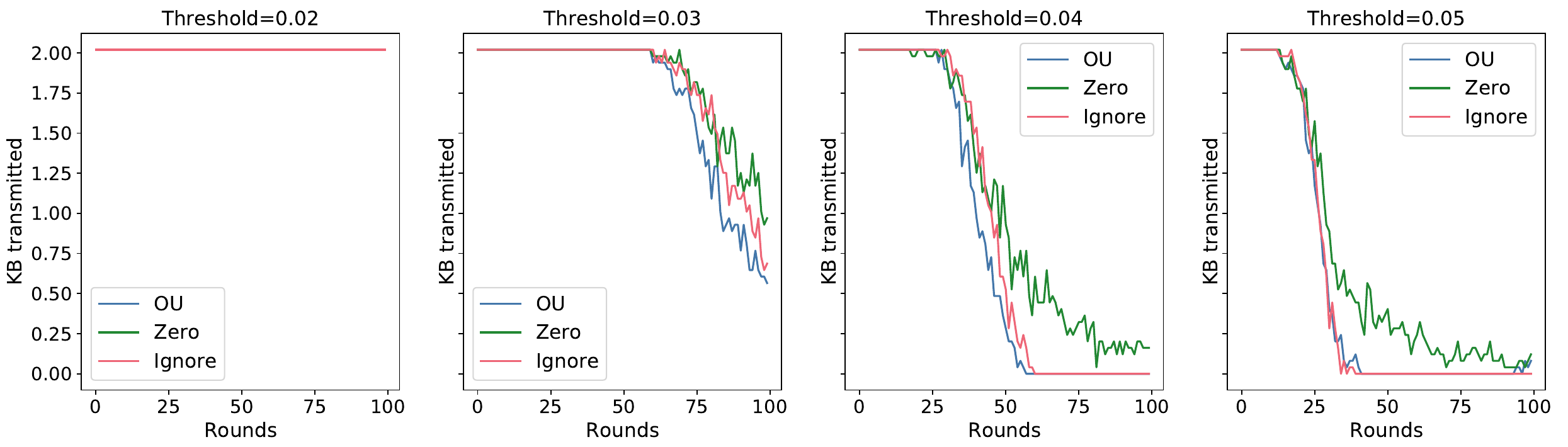}
    \caption{Per round communication of FedAvg with a fixed threshold communication strategy for varied thereshold values. Small thresholds fail to reduce communication, while high thresholds completely stop communication as training progresses thus leading to bad models.}
    \label{fig:fixed_threshold_communication}
\end{figure*}

We study the described FL systems in regards to the following main aspects of the client sampling problem: (i) communication efficiency vs. accuracy achieved by a client selection strategy; and (ii) the effects of different approaches to dealing with the clients that did not communicate their model updates to the server.

We first test our proposed client selection strategy and compare it with two baselines. In particular, we test our adaptive thresholding as well as its simpler variant wherein the threshold is fixed, and compare the results with those of the following baselines: (i) an approach where a fraction $q$ of $N$ clients is randomly dropped, with $q$ chosen to ensure a fair comparison with our thresholding schemes (specifically, $q$ is set equal to the average number of communicating clients per round in the thresholding scheme achieving the highest accuracy); and (ii) a non-restricted communication scheme where all clients communicate their model updates to the server. 


Then we turn our attention to the policy for dealing with missing updates and consider two simple alternatives: (i) \texttt{zero} strategy, where the server replaces missing updates with zeros (essentially assuming that the model of a client which did not send its update is identical to the latest global model); and (ii) \texttt{ignore}, where the server averages only the updates it received. For notational convenience, we refer to our strategy for predicting missing updates as \texttt{OU}. 

Note that in the experiments with EMNIST and Shakespeare datasets we fix hyperparameters as in the prior work \cite{reddi2020adaptive}, selecting $N=10$ clients at random in each round.  All the models are trained for 100 rounds.

\subsubsection{Fixed threshold} \label{sec:fixed_threshold}

As described in Section~3, our proposed thresholding strategy relies on varying the value of the threshold according to the mean and standard deviation of the norms of the model updates. It is worth considering if a simpler scheme employing a fixed threshold might suffice. First, note that for the fixed threshold strategy OU client selection outperforms competing methods. Specifically, we observe in two right-most plots in \fref{fig:fixed_threshold_accuracy} that \texttt{zero} and \texttt{ignore} strategies stagnate and tend to converge slower than the \texttt{OU} strategy. However, using a fixed threshold fails to provide accurate yet communication-efficient FL systems. First, treating threshold as a hyperparameter adds a layer of complexity to the system design problem since different values of the threshold may lead to very different results, as observed in \fref{fig:fixed_threshold_accuracy}. Tuning the threshold would go against the objective of reducing communication since a large number of rounds might be needed to facilitate the tuning. Finally, we observe that fixing the threshold to small values fails to reduce communication (left-most plot in \fref{fig:fixed_threshold_communication}) while setting it to large values leads to inaccurate models (right-most plot in \fref{fig:fixed_threshold_accuracy}).

\subsubsection{Evaluation of the system end-to-end}

We now evaluate our adaptive thresholding with OU estimation and compare it with the random selection strategy. In both contexts, random and thresholding,  we perform \texttt{OU}, \texttt{zero}, and \texttt{ignore} estimation strategies, both on EMNIST and Shakespeare datasets. We use a full communication strategy as a baseline for accuracy. 

\begin{figure*}[]
\centering
\begin{subfigure}{ 0.4\textwidth}
    \centering
    \includegraphics[width = \textwidth]{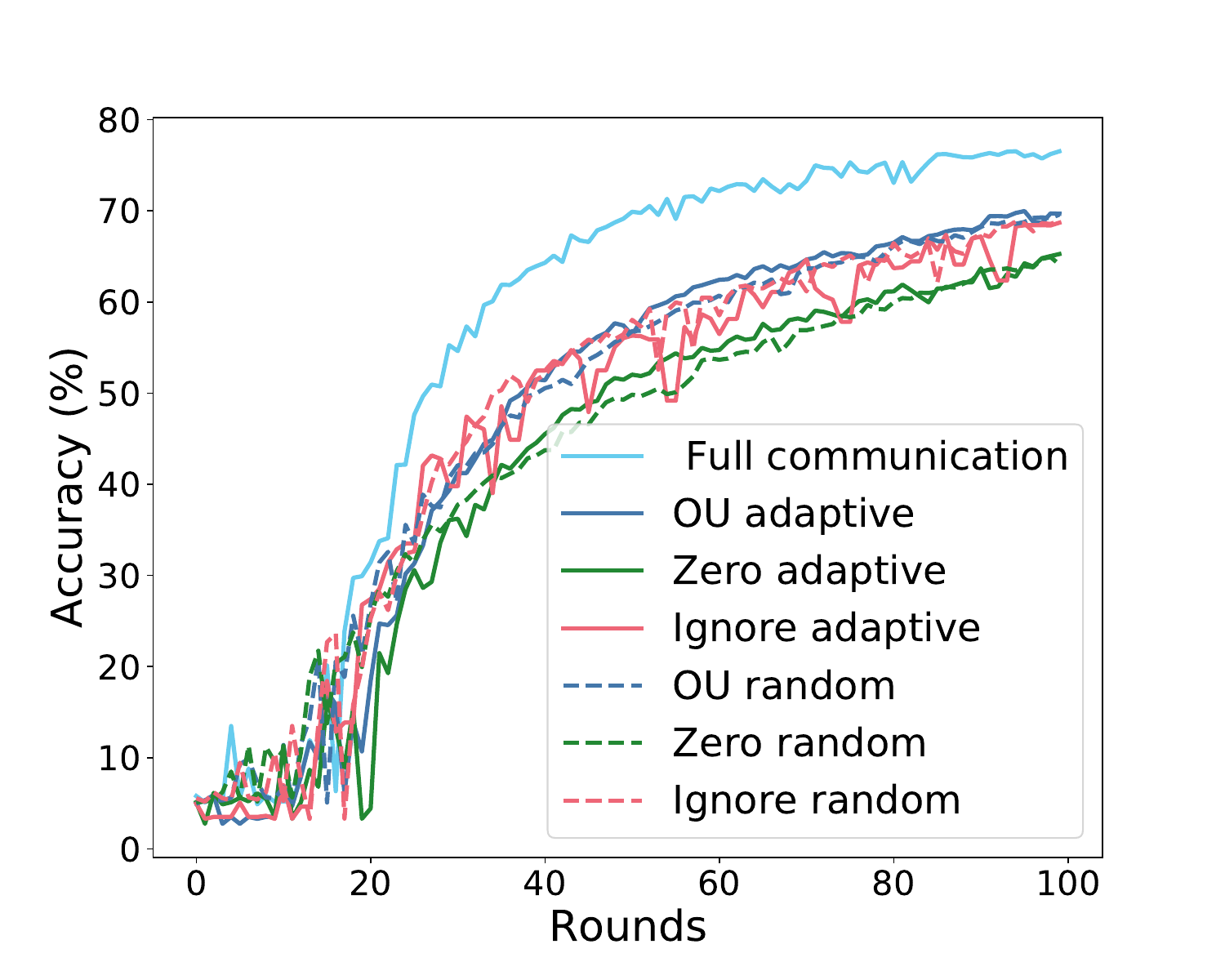}
    \caption{Test accuracy of EMNIST CNN.}
    \label{fig:accuracy_mnist}
\end{subfigure}
\begin{subfigure}{ 0.43\textwidth}
    \centering
 \includegraphics[width = \textwidth]{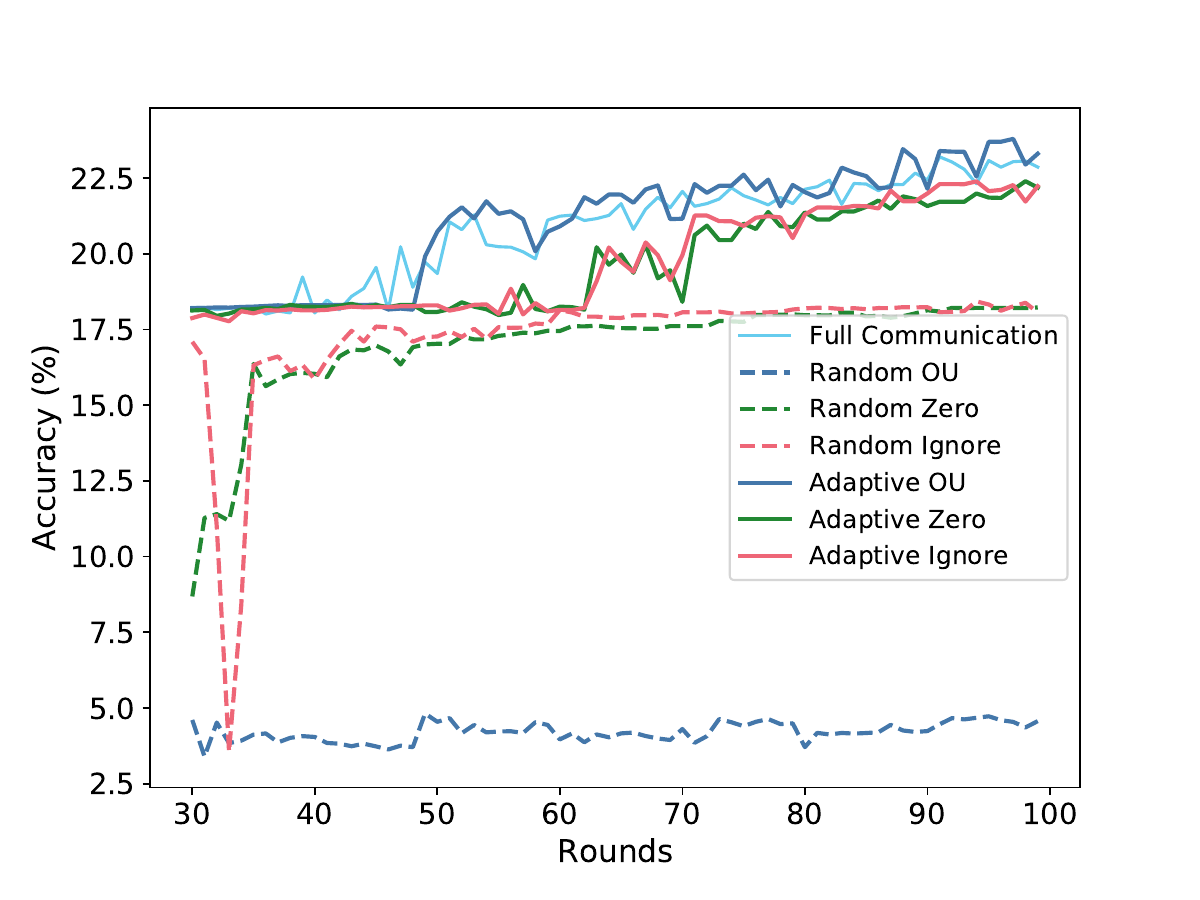}
 \caption{Test accuracy of Shakespeare RNN. }
    \label{fig:accuracy_shake}
\end{subfigure}
\caption{Model accuracy on synthetic logistic regression, EMNIST, and Shakespeare. The OU client sampling strategy consistently outperforms the competing methods.}
\label{fig:accuracies}
\end{figure*}

\paragraph{Accuracy.} In \fref{fig:accuracies}, we plot the test accuracy progression during training for all combinations of random/adaptive selection and the three estimation strategies, namely \texttt{OU, zero,} and \texttt{ignore}, both on EMNIST and Shakespeare datasets. We observe that the OU estimation is consistently better than the other strategies. Ignoring updates also achieves high accuracy, although it experiences a significantly more unstable convergence. Finally, \texttt{zero} strategy achieves the worst performance. We notice that the approaches which take into account missing updates and replace them with an estimate, \texttt{OU} and \texttt{zero}, help smooth the training process. However, by zeroing out clients' updates convergence is considerably slowed down on both datasets. On the other hand, \texttt{OU} is able to incorporate missing updates without rendering the convergence unstable. In both cases, the \texttt{OU} strategy achieves the best final accuracy; for the task involving Shakespeare dataset, \texttt{OU} even achieves better accuracy than the full-communication baseline scheme. 

\begin{figure*}[]
\centering
\begin{subfigure}{ 0.4\textwidth}
    \centering
    \includegraphics[width = 0.9\textwidth]{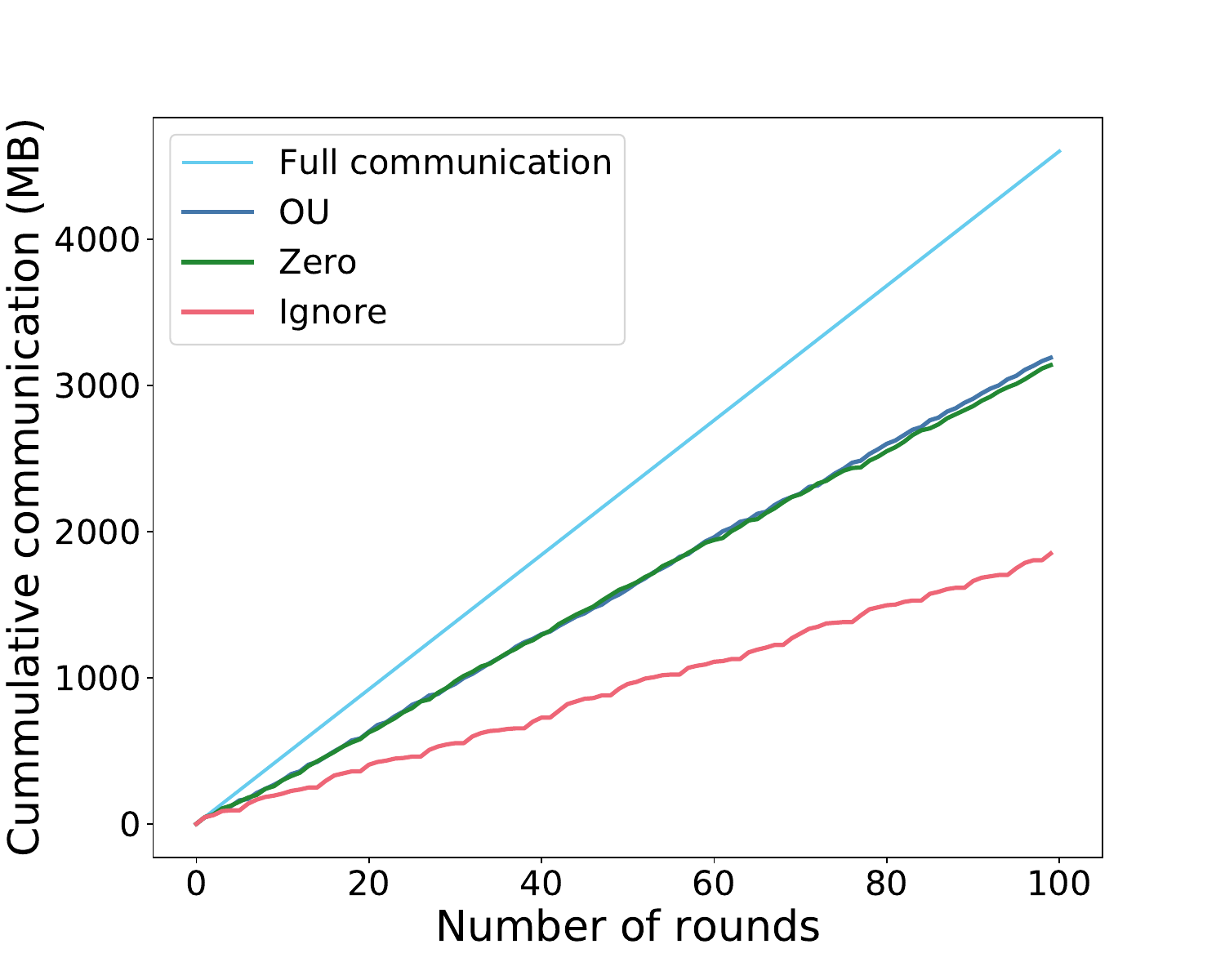}
    \caption{EMNIST CNN.}
    \label{fig:comm_mnist}
\end{subfigure}
\begin{subfigure}{ 0.4\textwidth}
    \centering
 \includegraphics[width = 0.9\textwidth]{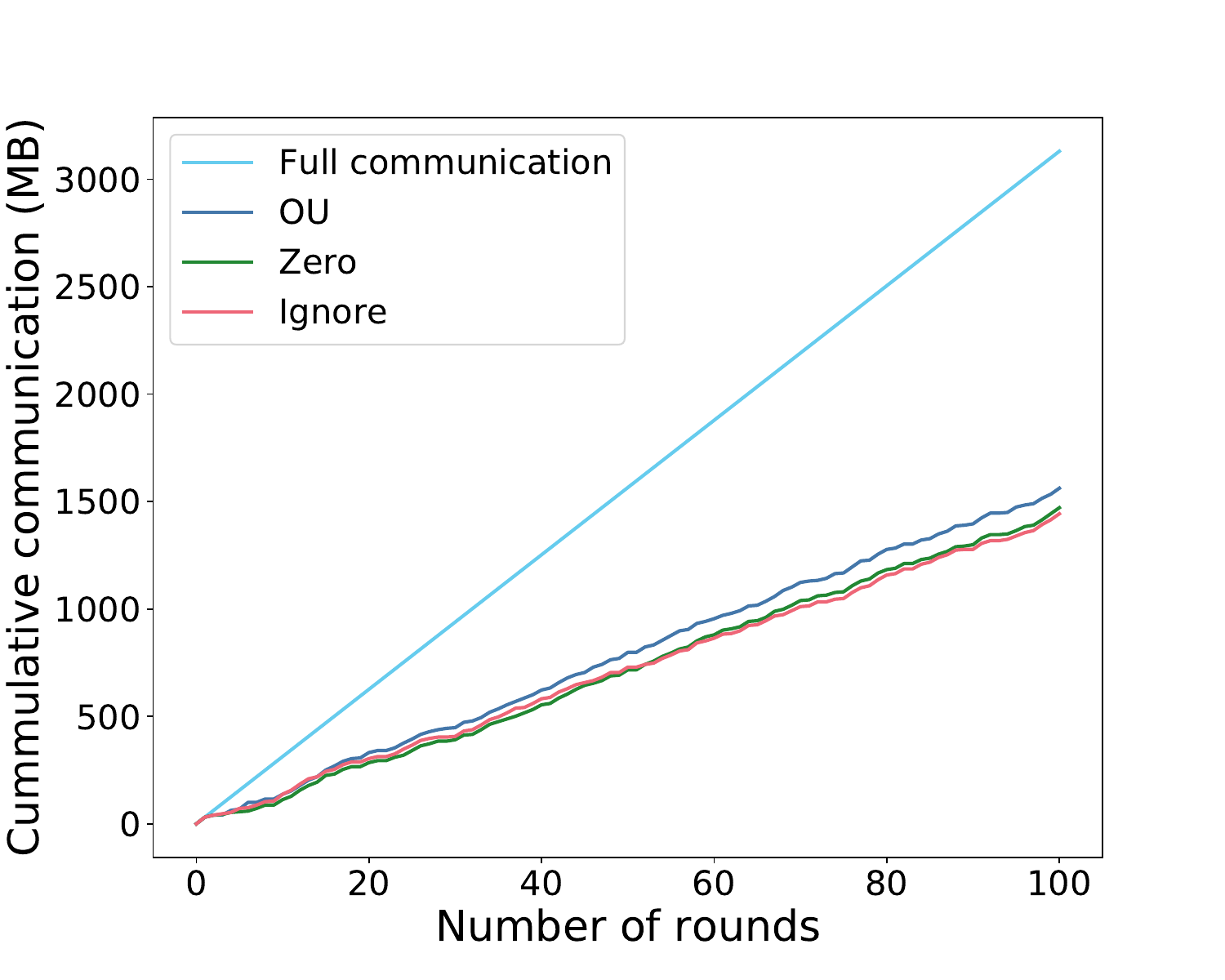}
 \caption{Shakespeare RNN. }
    \label{fig:comm_shake}
\end{subfigure}
\caption{Cumulative communication of each system during training measured in total KBs sent. \texttt{Ignore} strategy saves much more communication in the case of EMNIST CNN but does so at the expense of accuracy.}
\label{fig:cummulative_comm}
\end{figure*}

\paragraph{Communication savings. } 
As shown in \fref{fig:cummulative_comm}, our thresholding with the \texttt{OU} sampling and estimation requires smaller amount of communication to achieve accuracy comparable to the baseline (i.e., to the scheme using updates of all clients). The \texttt{zero} estimation strategy has similar communication savings to \texttt{OU} in all cases, but with slower convergence rates and inferior final accuracy. Among all strategies, the \texttt{ignore} strategy achieves the highest communication savings but is ultimately not capable of matching the accuracy of the \texttt{OU} method. This lower communication rate of the \texttt{ignore} strategy is due to the fact that by ignoring certain clients, they will continue to have high norm value in the subsequent iterations, which further inflates the threshold and leads to even fewer clients in the next round. We further elaborate on this in the supplementary (\sref{sec:threshold}).

While randomly dropping clients trivially achieves low communication rate (by definition equivalent to its adaptive estimation counterpart in each case), the random client selection scheme suffers from a deterioration in accuracy in heterogeneous and non-convex settings. 
In non-homogeneous, non-convex settings, estimating the true gradient from a subset of clients is much harder, and thus randomly dropping clients slows down the convergence. It is also in part consequence of dropping too many clients near the optimum where gradient norms might become smaller. Our adaptive thresholding strategy overcomes aforementioned problems by changing the threshold in each round based on the clients' update norms. In the random setting, we observe that ignoring or zeroing updates is better, which accentuates the coupling of our thresholding and estimation strategies: by thresholding and using the MSE estimator we optimally incorporate the missing clients.

Our results demonstrate that the level of accuracy in FL can be maintained, or even outperformed, while reducing communication by using a smaller number of clients; however, these clients have to be carefully selected, and their updates have to be carefully incorporated in the new model. 

The results in \tref{tab:results_emnist_shake}
suggest that while there is no single method which is uniformly superior in exploring accuracy-communication trade-offs, the thresholding strategies are an efficient way of subselecting clients without a significant deterioration of accuracy.

\begin{table*}[]
\caption{Accuracy and communication cost of communication-reduction strategies in EMNIST and Shakespeare dataset experiment.}
\label{tab:results_emnist_shake}
\hskip-2.0cm
\begin{tabular}{lllllll}
\toprule
                         & \multicolumn{2}{l}{\textbf{Accuracy}}                                        & \multicolumn{2}{l}{\textbf{Overall}}    & \multicolumn{2}{l}{\textbf{Communication}} \\
                         & \multicolumn{2}{l}{(\%)}                                                     & \multicolumn{2}{l}{\textbf{comm. (Gb)}} & \multicolumn{2}{l}{used (\%)}               \\ \midrule
Dataset                  & EMNIST                                & Shakespeare                          & EMNIST           & Shakespeare          & EMNIST            & Shakespeare            \\ \midrule
\textbf{Baseline}        & 76.51                                 & 22.86                                & 4.5              & 3.1                  & 100               & 100                    \\ \midrule
\textbf{Adaptive OU}     & { \textbf{69.67}} & { \textbf{23.3}} & 3.15             & 1.53                 & 70                & 49.9                   \\
\textbf{Adaptive zero}   & 65.25                                 & 22.18                                & 3.01             & 1.44                 & 68.9              & 47                     \\
\textbf{Adaptive ignore} & 68.69                                 & 22.23                                & 1.85             & 1.41                 & 41.1              & 46.1                   \\
\textbf{Random OU}       & 69.58                                 & 4.56                                 & 3.16             & 1.46                 & 70.34             & 47.9                   \\
\textbf{Random zero}     & 64.11                                 & 18.22                                & 3.17             & 1.45                 & 70.6              & 47.3                   \\
\textbf{Random ignore}   & 68.91                                 & 18.06                                & 1.75             & 1.39                 & 38.9              & 45.4  \\            \bottomrule    
\end{tabular}
\end{table*}
\section{Conclusion}
\label{sec:conclusion}

In this paper, we propose a novel approach to reducing communication rates in FL by judiciously subselecting clients instead of relying on traditional compression strategies. A parallel between OU processes and SGD suggests strategies for identifying clients whose updates are informative and therefore should be communicated to the server. Furthermore, utilizing this connection leads to an estimator for missing client model updates that can be calculated at the server using the previous updates. This estimation helps maintain and even improve the accuracy of the baseline scheme, while cutting communication by up to 50\%. Experimental results demonstrate efficacy of the proposed methods in various settings. Moreover, our approach can be combined with compression strategies to lower the communication rates even further. 

The proposed client selection protocol based on thresholding is theoretically justified by the existing results on optimal OU process sampling. 
Future work involves combining the methods proposed in this paper with the techniques that attempt to reduce communication burden by sparsifying transmitted information.




\bibliography{references}
\bibliographystyle{abbrvnat}

\newpage
\appendix
\onecolumn
\section{Background details}
\subsection{Federated Averaging algorithm}
\label{sec:fedavg}
For convenience and completeness, we here provide \texttt{FedAvg}, the baseline federated learning algorithm proposed in \cite{mcmahan17a}.

\begin{algorithm}

    \KwIn{$K$ clients, $B$ is the local minibatch size,$E$ is the number of local epochs, and $\eta$ is the learning rate. }
    \KwOut{ Global model $w$ }
    initialize $w_0$\;
         \For{t=1,2,...}{ 
            $m \gets \max(CK,1)$ \;
            $S_t \gets$ random set of $m$ clients \;
             \For{ $k\in S_t$ in parallel}{
                 $w^{k}_{t+1} \gets$ ClientUpdate($k,w_t$) \;
                }
            $w_{t+1} \gets \sum_{k=1}^K \frac{n_k}{n}w_{t+1}^k$ \;
            }
    \textbf{Clients execute:} \;
    ClientUpdate($k,w$) : Run on client $k$ \;
        $\mathcal{B} \gets $ split data $\mathcal{P}_k$ into batches of size $B$ \;
        \For{$i =1,...,E$}{
            \For{$b\in \mathcal{B}$}{
                $w \gets w-\eta \nabla \ell(w,b)$ \;
           }
        }
       \Return{$w$ to server} \;
 \caption{\texttt{FederatedAveraging} \cite{mcmahan17a}}
 \label{alg:FedAvg}
\end{algorithm}

\section{Experimental Details}
\subsection{Numerical illustration of SGD: Aptness of the OU process model} \label{sec:numerical_illustration}

To illustrate the aptness of using OU process for modeling sequences of weights/parameters during SGD, we performed a simple experiment on the Fashion MNIST dataset \cite{xiao2017/online}. This dataset consists of 60,000 training images and 10,000 test examples; each sample is a 28x28 image, belonging to one out of 10 categories. No user or id information is available. 

We train a feed forward neural network with a single hidden layer consisting of 128 units, ReLU activation, and a softmax output layer. We train the model for 5 epochs with the batch size set to 100, for a final test accuracy of 84\%, when the model saturates. 

\begin{figure}[]
    \centering
    \includegraphics[width = 0.5 \textwidth]{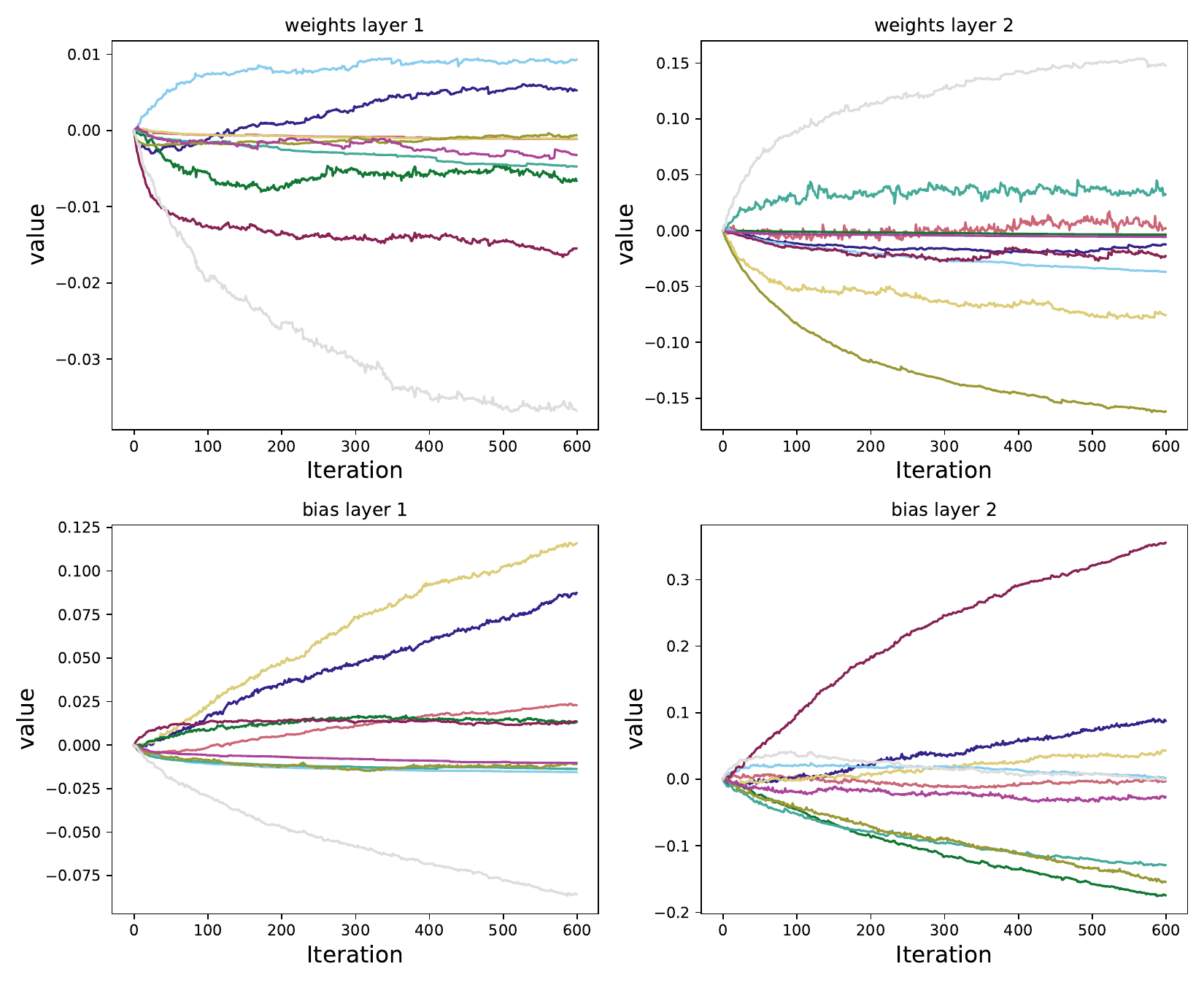}
    \caption{Parameter drift when retuning the model with new data. The weights follow trajectories reminiscent of the OU process sample paths.}
    \label{fig:modelDrift}
\end{figure}

In \fref{fig:modelDrift} we show the trajectories of individual weights $w^i$ randomly selected from each layer (we plot the trajectories after centering by setting $w^i_0 = 0$). Similar to sample paths of an OU process, all trajectories in Fig 1 ultimately deviate towards a mean where they stabilize.

\subsection{Datasets}

The datasets used for benchmarking and model validation experiments are: (i) EMNIST dataset, a reprocessed version of the original MNIST dataset where each image is linked to its original writer, providing a non-i.i.d. natural distribution and allowing us to emulate an FL setting; and (ii) Shakespeare dataset, a language modelling dataset with questions and answers collected from users. Size of the datasets is summarized in the table below.

\begin{table}
\caption{Datasets}
\centering
\begin{tabular}{@{}lcc@{}}
\toprule
Dataset       & Users & Samples \\ \midrule
EMNIST        & 3.4 K & 60 K    \\
Shakespeare & 715 & 16 K   \\ \bottomrule
\end{tabular}
\end{table}

\subsection{Models}
\label{sec:app:architectures}

The ML architectures that we used are presented in Tables~1-2 below.

\paragraph{EMNIST CR}
We train a convolutional neural network with the following architecture:  two 5x5 convolution layers, with 32 and 64 channels respectively, interleaved with 2x2 max-pooling, a dense layer with 512 neurons with ReLU activation and a 10 unit softmax output layer. The network has a total of 1,663,370 parameters. 
 At each round, 50 clients are uniformly selected to update the model. Each client locally trains for $E=20$ epochs using stochastic gradient descent (SGD) with a batch size of $B=10$.


\begin{table}[]
\centering
\caption{ EMNIST digit recognition convolutional model architecture}

\begin{tabular}{@{}lcccl@{}}
\toprule
\multicolumn{1}{c}{\textbf{Layer}} & \textbf{Output} & \textbf{\begin{tabular}[c]{@{}c@{}}\# Trainable \\ parameters\end{tabular}} & \textbf{Activation} & \textbf{Hyperparameters} \\ \midrule
Input                              & (28,28,1)       &                                                                             &                     &                          \\
Conv2d                             & (26,26,32)      & 320                                                                        & ReLU                & kernel size = 3; strides(1,1)          \\
Conv2d                             & (24,24,64)      &   18496                                                                  &                     & kernel size = 3; strides(1,1)     \\
MaxPool2d                          & (12,12,64)     &  0                                                                             &                     & pool size= (2, 2)        \\
Dropout                             & (12,12,64)    & 0                                                                             &                               & p=0.25\\
Flatten                            & 9216           &    0                                                                         &                     &                          \\
Dense                              & 128             & 1179776                                                                     &                 &                          \\
Dropout                             & 128           & 0                                                                             &                               & p=0.5\\
Dense                              & 62              & 7998                                                                        & Softmax             &                          \\ \midrule
\multicolumn{1}{c}{\textbf{Total}} &                 &  1,206,590                                                           &                     &                          \\ \bottomrule
\end{tabular}
\end{table}

\paragraph{Shakespeare}

We train a recursive neural network for the next character prediction that first embeds characters into an 8-dimensional space, followed by 2 LSTMs and finally a dense layer. The architecture is presented in Table \ref{tab:ShakespeareArchit}.

\begin{table}[]
\caption{ Shakespeare next character prediction model architecture}
\label{tab:ShakespeareArchit}
\centering
\begin{tabular}{@{}lccc@{}}
\toprule
\multicolumn{1}{c}{\textbf{Layer}} & \textbf{Output} & \textbf{\begin{tabular}[c]{@{}c@{}}\# Trainable \\ parameters\end{tabular}} & \textbf{Activation}  \\ \midrule
Input                              & 80              & \multicolumn{1}{l}{}                                                        & \multicolumn{1}{l}{} \\
Embedding                          & (80,8)         & 720                                                                      &                      \\
LSTM                               & (80,256)        & 271360                                                                     &                      \\
LSTM                               & (80,256)         & 525312                                                                    &                      \\
Dense                              & (80,90)      & 23130                                                                     & Softmax              \\ \midrule
\multicolumn{1}{c}{\textbf{Total}} &                 & 4,050,748                                                                 &                      \\ \bottomrule
\end{tabular}
\end{table}


\subsection{Number of clients} 

Recall that when a new round starts, the server samples a fixed predetermined number $N$ of clients. As one would expect, increasing $N$ leads to improvement of accuracy of all the methods considered. We test how the performance vary with $N$ and show that our method consistently outperforms all the other strategies (see Table 7). 

\begin{table*}[]
\caption{Accuracy (\%) on Shakespeare after $120$ rounds, with varying number of initial clients $N$}
\label{tab:num_clients_accuracy}
\hskip-2.0cm
\small
\begin{tabular}{@{}llllllll@{}}
\toprule
Method & Ada. OU & Ada. ignore & Ada. zero & Random OU & Random ignore & Random zero & Full comm. \\ \midrule
N=10                 & \textbf{23.3 }   & 22.22       & 22.18     &    4.98       & 18.05         & 18.22       & 22.8               \\
N=20                 & \textbf{23.96 }  & 23.37       & 23.06     &     0.0      & 20.57         & 18.26       & 23.55              \\
N=50                 & \textbf{25.17 }  & 24.31       & 24.78     & 24.51     & 21.24         & 21.03       &            26.63        \\ \bottomrule
\end{tabular}
\end{table*}

\begin{table}[]
\caption{Communication rate (\%) on Shakespeare with varying number of initial clients $N$}
\label{tab:num_clients_communication}
\centering
\begin{tabular}{@{}llll@{}}
\toprule
Method  & Ada. OU       & Ada. ignore    & Ada. zero \\ \midrule
N=10                 & \textbf{46.6} & 46.02          & 47.3      \\
N=20                 & 47.43         & \textbf{47.14} & 47.98     \\
N=50                 & 48.95         & \textbf{48.93} & 49.13     \\ \bottomrule
\end{tabular}
\end{table}

\subsection{Comparing thresholds across strategies}
\label{sec:threshold}

Assume adaptive thresholding schemes where a threshold is re-computed in each round based on statistics of the norms of the updates of participating clients (see line 12 in \aref{alg:all_system}); we empirically study thresholds generated by different strategies and the resulting communication rates. As expected, a strategy has an effect on the distribution of the magnitudes of the updates, leading to different thresholds and varied communication savings. In particular, in \fref{fig:cummulative_comm} we observe that while \texttt{OU} and \texttt{zero} utilize similar levels of communication, \texttt{ignore} achieves more savings; this is particularly pronounced on EMNIST.

\begin{figure*}[]
\centering
\begin{subfigure}{ 0.4\textwidth}
    \centering
    \includegraphics[width = \textwidth]{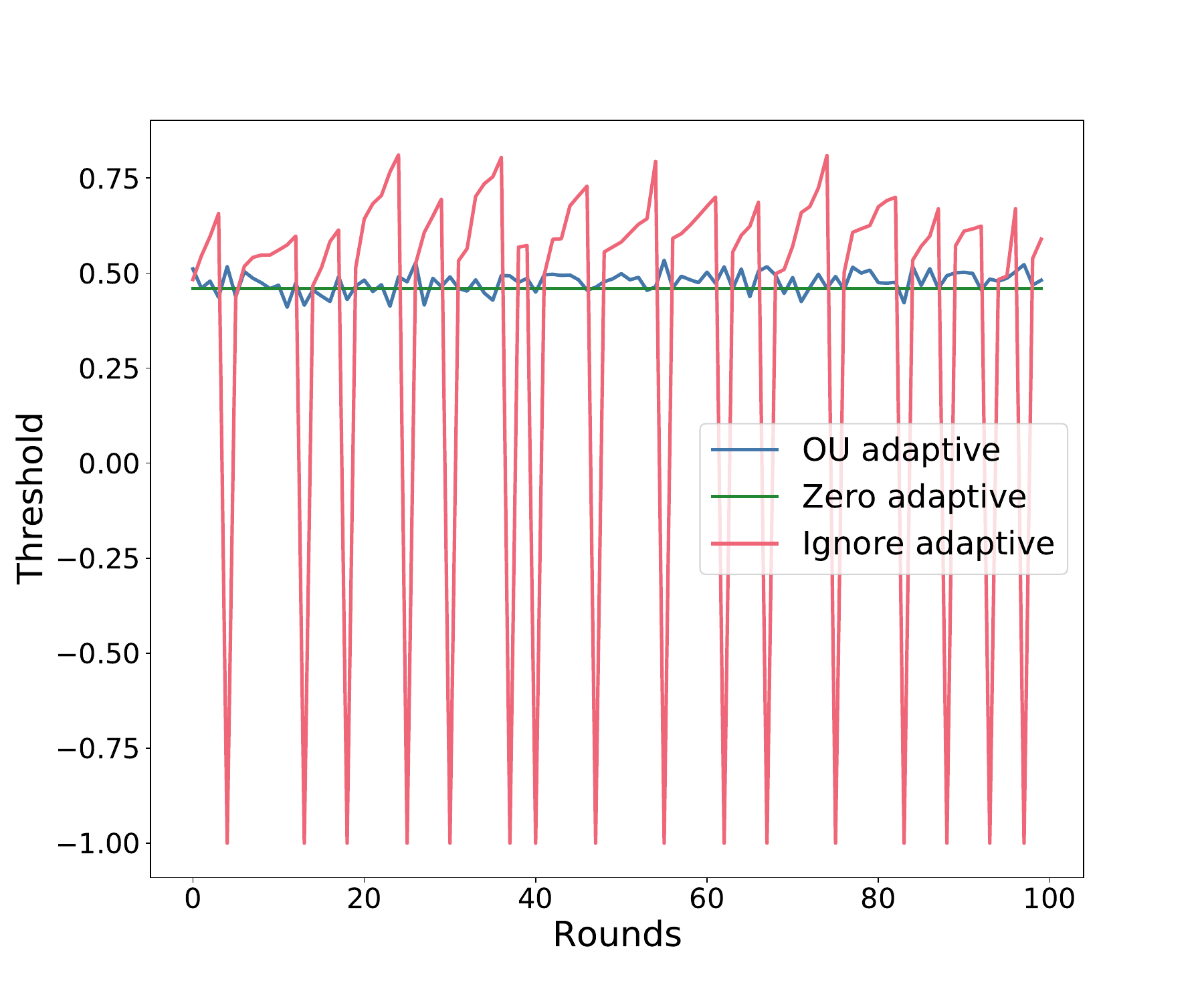}
    \caption{Threshold EMNIST CNN.}
    \label{fig:thresh_mnist}
\end{subfigure}
\begin{subfigure}{ 0.4\textwidth}
    \centering
 \includegraphics[width =  \textwidth]{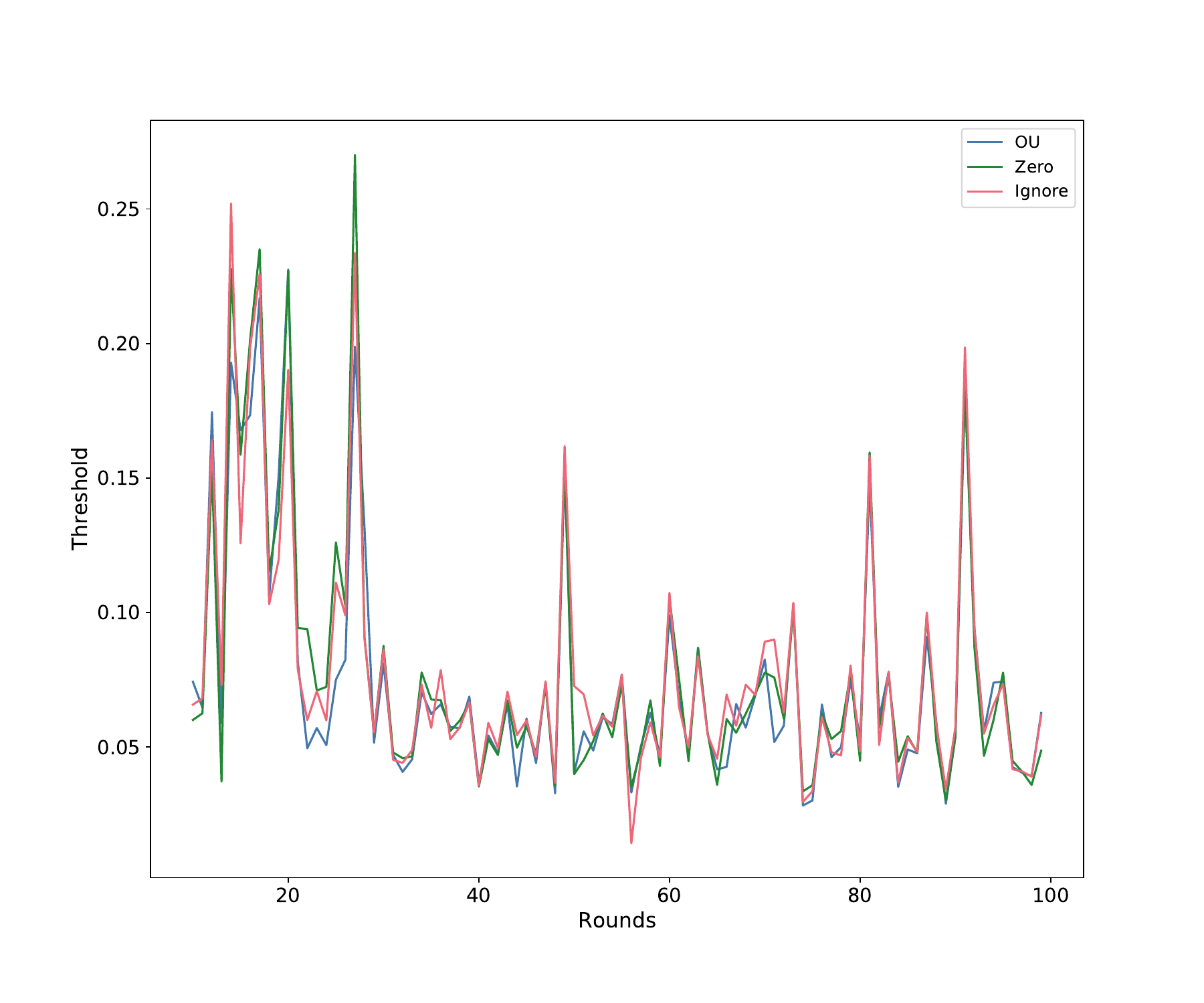}
 \caption{Threshold of Shakespeare RNN. }
    \label{fig:thresh_shake}
\end{subfigure}
\caption{Threshold values across rounds for different strategies.}
\label{fig:thresholds}
\end{figure*}

In \fref{fig:thresholds} we show the progression of thresholds for each strategy and in \tref{tab:thresh} we display the mean and variance of threshold across rounds. We observe the instability of \texttt{ignore}: by ignoring certain clients, the global model becomes inadequate for this group of clients; consequently, this estimation strategy may lead to larger future updates of those clients. The thresholds may take on larger values as well while, at the same time, experiencing an increased variance that may drive the threshold in some rounds to 0 (or to negative values which simply means that all clients transmit). \texttt{OU} and \texttt{zero} have a significantly more stable schedule of thresholds leading to smoother behavior of the magnitudes of model update norms; clearly, such benefits stem from  incorporating estimates of the missing client updates in the computation of the next global model.

\begin{table}[]
\centering
\caption{Mean threshold values across rounds.}
\label{tab:thresh}
\begin{tabular}{@{}lll@{}}
\toprule
       & EMNIST           & Shakespeare \\ \midrule
OU     & 0.476 $\pm$ 0.03 & 0.746       \\
Zero   & 0.459 $\pm$ 0.0  & 0.776       \\
Ignore & 0.515 $\pm$ 0.24 & 0.775       \\ \bottomrule
\end{tabular}
\end{table}

\end{document}